\documentclass[10pt,twocolumn,letterpaper]{article}
\usepackage{xcolor}
\definecolor{wzc}{RGB}{255,182,193} 
\usepackage{pifont}
\usepackage{tabularx}

\usepackage{colortbl}  
\definecolor{DarkGreen}{RGB}{0,100,0}
\definecolor{DarkRed}{RGB}{139,0,0}
\usepackage[review]{iccv}      

\definecolor{iccvblue}{rgb}{0.21,0.49,0.74}
\usepackage{multirow}
\usepackage[pagebackref,breaklinks,colorlinks,allcolors=iccvblue]{hyperref}

\title{
IPFormer-VideoLLM: Enhancing Multi-modal Video Understanding  for Multi-shot  Scenes}

\author{Yujia Liang \quad Jile Jiao\quad  Xuetao Feng\quad  Zixuan Ye \quad  Yuan Wang\quad  Zhicheng Wang\thanks{Corresponding author.}\\
School of AIA, Huazhong University of Science and Technology\\
Deepeleph Intelligent Technology\\
JD Explore Academy\\
Department of Electronic Engineering, Tsinghua University\\
{\tt\small 
 \{yjl, zhicheng\_wang\}@hust.edu.cn}
}
\begin{document}
\maketitle
\begin{abstract}

Video Large Language Models (VideoLLMs) have demonstrated remarkable understanding capabilities, but are found struggling to 
tackle multi-shot scenarios, \textit{e.g.}, video clips with varying camera angles or scene changes. 
This challenge can render failures such as instance identity forgetting and key frame negligence. In this work, we first attribute the challenge to the lack of multi-shot annotations among existing datasets and therefore we introduce a new dataset termed \textbf{MultiClip-Bench}, featuring dense descriptions and instruction-based question-answering pairs tailored for multi-shot scenarios. We empirically find that the training set significantly boosts the multi-shot performance, while the testing benchmark provides a reliable measure of 
the model 
capability in multi-shot scenarios. 
By further analyzing and discovering that current models only encode instance features in a discrete or lossy manner, at the risk of missing identity information, we then contribute a new model 
\textbf{IPFormer-VideoLLM}. 
Its key idea is the injection of instance-level features as instance prompts through an efficient attention-based connector. This allows for the aggregation of instance-specific information across scenes. Experiments demonstrate that our proposed dataset and model not only enhance the multi-scene video understanding significantly, but also offer distinct advantages across various video benchmarks. The code and model will be available.
\end{abstract}    
\begin{figure}
	\centering
	\includegraphics[width=0.95\linewidth]{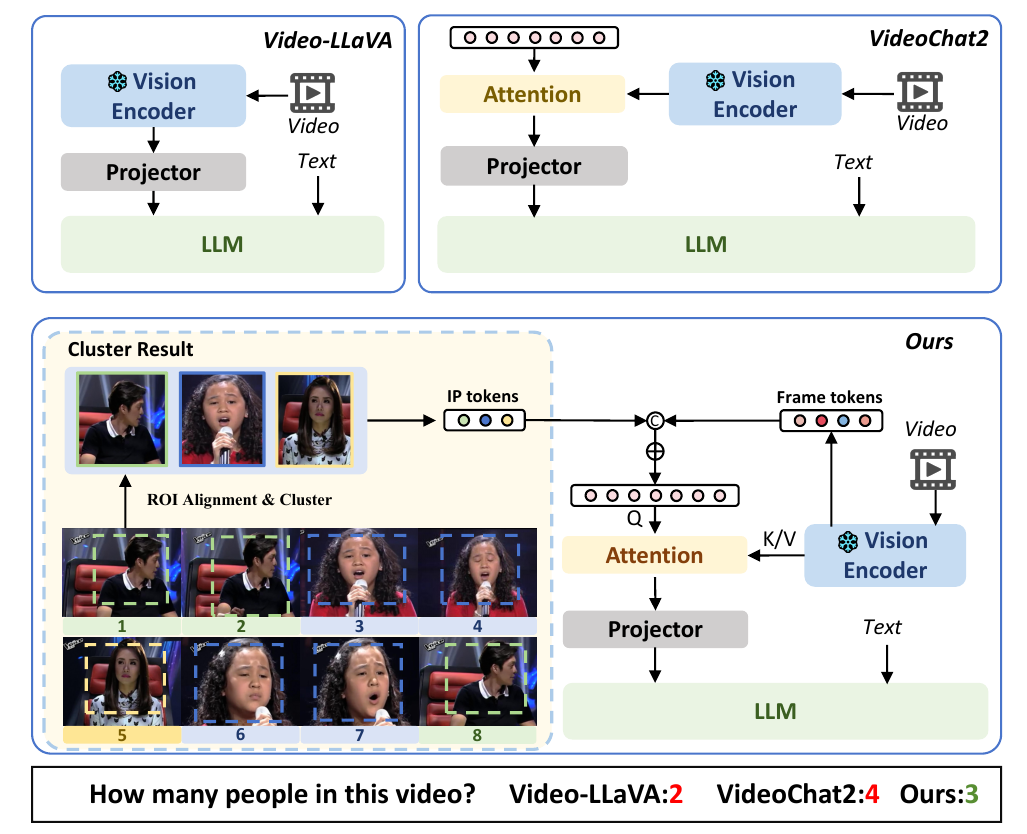}
\caption{\textbf{Multi-shot Transitions} Video-LLaVA uses full projection, while VideoChat2 reduces video tokens via attention mechanisms. In scenes with discontinuous character appearances (multi-scene transitions), existing models often fail to distinguish characters, leading to counting errors. Our method compresses tokens and uses detectors and clustering to generate instance prompts, to guide attention fusion across scenes and enhance accuracy.
}	\label{fig:1}
\vspace{-13pt}
\end{figure}

\section{Introduction}
Large Multi-modal Models (LMMs) ~\cite{alayrac2022flamingo,driess2023palm,huang2023language,li2023blip} built by connecting Large Language Models (LLMs) as the brain with specialized encoders, 
have yielded remarkable progress in tasks involving multiple modalities.
In the domain of video,
recent Video Large Language Models (VideoLLMs)~\cite{li2023videochat,li2024mvbench,liu2024mmbench,bai2025qwen2,li2024llava,wang2022internvideo} 
have customized domain-specific model architectures~\cite{team2024gemini,li2023videochat,cheng2024videollama,qu2024ts} and datasets/benchmarks~\cite{maaz2024videogpt+,krishna2017dense,wang2023internvid}, 
significantly advancing many video understanding problems such as long video understanding~\cite{liu2024st,song2024moviechat}, fine-grained understanding~\cite{wang2022internvideo,li2024llava}, and temporal localization~\cite{chen2024timemarker,wang2024grounded,patraucean2023perception,li2023seed}. 

Although current VideoLLMs perform 
well in 
most scenarios, 
they exhibit notable deficiencies in multi-shot scenes, \textit{e.g.}, 
video clips captured from various angles of view or different scenes. Compared to single-shot videos, multi-shot ones involve more visual clues and more 
complex object relationships, requiring 
advanced information extraction and instance-level comprehension capabilities. Fig.~\ref{fig:1} illustrates a failure case in a multi-shot scenario. Given 
video clips from different cameras depicting 
the same talent show, the models are asked to count the number of people who have ever appeared. Surprisingly, for this seemingly simple question, Video-LLaVA~\cite{lin2023video} and VideoChat2~\cite{li2024mvbench} answered $2$ and $4$ respectively, while the ground truth is $3$. 

Why do existing VideoLLMs fail in such an easy case? We attribute the failure to two 
key factors: data and model limitations. First, an examination of commonly used datasets reveals a scarcity of training data for multi-shot scenarios. Current annotations 
primarily focus on single-shot videos, with monotonous camera changes, slow object movement, and few scene transitions. In contrast, the descriptive quality of high-dynamic multi-shot videos remains substandard. Those multi-shot annotations tend to provide only brief overviews of the video, while lacking details and depth. 
Moreover, on the test end, the widely used benchmarks \cite{wang2023internvid,krishna2017dense,maaz2024videogpt+,li2024mvbench,fu2024video} can hardly reflect the ability for multi-shot video understanding due to the absence of relevant test cases. 

To fill this gap, we first introduce a multimodal video understanding benchmark called \textbf{MultiClip-Bench}. The creation of MultiClip-Bench involves both video annotation and question-answer (QA) formulation. For annotation, we segment the video into key frames, annotate each key frame with character IDs, and label the video with dense captions and character information.
In the design of QA pairs, we also focus on the challenges of multi-shot scenarios, particularly on cases where characters appear discontinuously across multiple scenes or have brief appearances, and generate a large number of QA pairs accordingly.
Finally, we manually selected a subset of $2,750$ pairs to constitute the test set, with answers verified, refined, and categorized to form the benchmark.

Although significant improvement is observed when using the MultiClip training set, we found several typical failure cases of current models when handling multi-shot videos. Two significant deficiencies of current VideoLLMs are identified: i) the lack of instance consistency preservation during scene transitions; and ii) the occlusion of information about briefly appearing characters caused by large data volumes in the video. To address these challenges, 
we propose a new model, \textbf{IPFormer-VideoLLM}, which utilizes \textit{instance prompts} to guide attention for more effective instance feature preservation. Specifically, we sample and aggregate features from bounding-box regions to form instance features, and inject them as visual prompts into the query to guide attention. Besides significant performance gains, it reduces tokens to less than $10\%$ of the baseline, cuts training time by $75\%$, and boosts inference speed. 

Experimental results demonstrate that our proposed model performs strongly on both existing benchmarks and our MultiClip-Bench, with further performance improvements observed when utilizing our proposed training dataset. Specifically, compared to the baseline model Video-LLaVA~\cite{lin2023video}, our model achieves a significant improvement of \textcolor{DarkGreen}{\textbf{2.3\%}}  on MultiClip-Bench. With the support of our training dataset, the performance on MultiClip-Bench increases by \textcolor{DarkGreen}{\textbf{8.5\%}}. Noticeable performance gains are also observed on other datasets. For example, datasets such as NExT-QA~\cite{xiao2021next} and IntentQA~\cite{li2023intentqa} exhibit particularly remarkable enhancements. 
This further validates the significant value of our dataset in addressing the common limitations of multi-shot video understanding.
\section{Related Work}
\label{sec:Related Work}

\subsection{Video Benchmark}
Video benchmark evaluations can be categorized into comprehensive assessment sets and task-specific assessment sets. Modern frameworks include MMBench~\cite{liu2024mmbench}, Video-MME~\cite{fu2024video}, and MVBench~\cite{li2024mvbench},  among others. Specifically, Video-MME enhances data diversity through domain hierarchies (6 categories, 30 subclasses) and balanced video durations. MVBench provides a comprehensive evaluation by categorizing tasks into 20 types, facilitating holistic assessment. Task-specialized benchmarks target specific capabilities: Perception Test~\cite{patraucean2023perception} evaluates spatiotemporal reasoning, FunQA~\cite{xie2024funqa} tests models with counter-intuitive scenarios, NExT-QA~\cite{xiao2021next} focuses on object interactions, and IntentQA~\cite{li2023intentqa} emphasizes understanding user intent. However, existing benchmarks primarily concentrate on single-shot or simple scene transitions, offering limited comprehension difficulty, and fail to fully address the unique challenges of multi-shot scenes, such as maintaining character consistency across discontinuous scenes, interpreting short-duration segments, and recognizing outlier events.

\subsection{Video Feature Alignment}
In MLLMs, image or video tokens are flattened and projected after feature extraction, and then concatenated with text before being fed into the LLM~\cite{lin2023video,liu2023visual,liu2024improved}. Although this approach is simple and effective, the large number of tokens, especially those from videos, significantly increases computational costs for the LLM. As MLLMs are increasingly deployed in real-time applications~\cite{ren2024timechat,song2024moviechat,bai2025qwen2,huang2024vtimellm}, enhancing inference efficiency and accommodating more frames have become critical research goals, particularly in multi-shot scenarios. Currently, reducing the number of tokens input to the LLM constitutes the primary strategy. Token compression techniques include downsampling through average pooling~\cite{xu2024pllava,bai2025qwen2,li2024llava}, fast-slow stream merging~\cite{xu2024slowfast,qu2024ts}, compressing videos into fixed-length representations using memory techniques~\cite{song2024moviechat}, and attention-based compression via Q-former~\cite{zhang2023video,bai2023qwen,li2023videochat}. However, crucial information can be lost during token compression, particularly in complex multi-shot scenes. To address this issue, we propose an efficient compression scheme specifically tailored for multi-scene applications. Our approach can accommodate a greater number of frames while effectively preserving essential information.

\begin{figure*}[!t]
	\centering
	\includegraphics[width=0.92\linewidth]{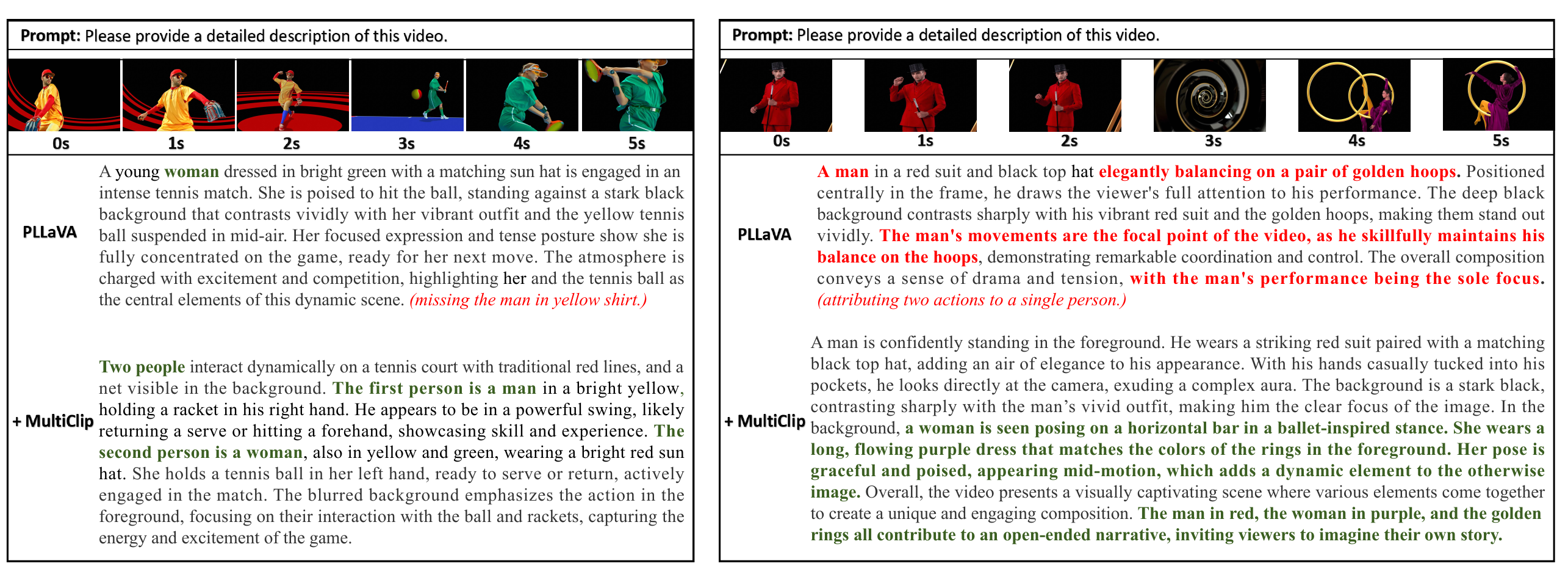}
\caption{\textbf{Challenges in Multi-shot Scenarios.} In multi-shot scenarios, Dense description capability of PLLaVA  falters. As shown on the left, the system misses some key frames, only recognizing the woman in green while failing to detect the man in yellow. In the example on the right, the system exhibits person confusion, incorrectly attributing the actions of two individuals to a single person.
}
	\label{fig:PLLAVA}
\end{figure*}
\section{MultiClip-Bench}
In this section, we first describe the process of filtering multi-shot videos and creating target video descriptions in \ref{3.1}. Next, we will introduce the generation of multi-shot related QA pairs in \ref{3.2}.


\subsection{Multi-Shot Description Generation}
\label{3.1}
\begin{figure*}
	\centering
	\includegraphics[width=0.9\linewidth]{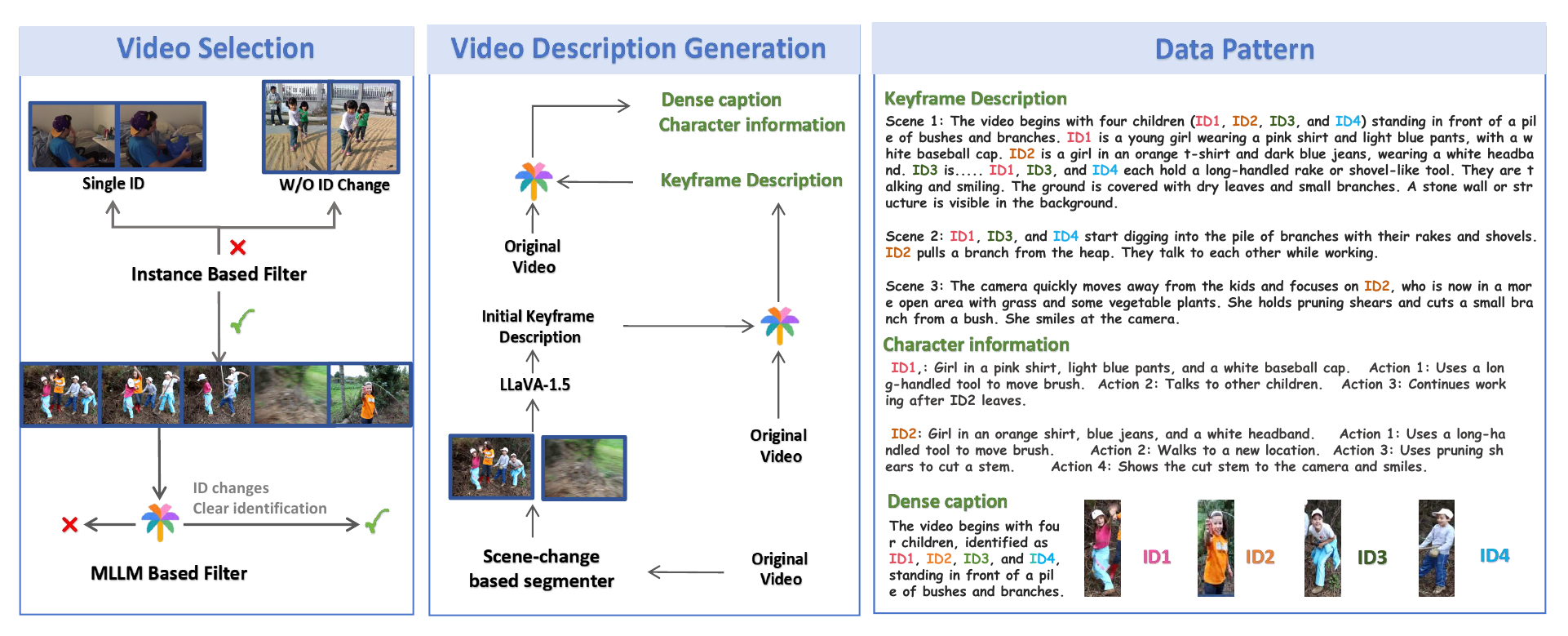}
\caption{\textbf{Description Annotations of MultiClip-Bench} First, we filter the target videos, and then use close-source large video models to generate the three key elements of video description: \textbf{keyframe description}, \textbf{character information}, and \textbf{dense caption}. 
	}
	\label{fig:2}
\end{figure*}

Some widely used video LLMs, such as PLLaVA~\cite{xu2024pllava}, claim to possess strong video understanding capabilities. However, in multi-shot videos (Figure \ref{fig:PLLAVA}), previous VideoLLMs exhibit notable drawbacks, struggling to recognize even simple scenes transitions. A major reason is the lack of tailored multi-shot video data. Motivated by this requirement, we propose MultiClip-Bench, the first multi-shot video dataset, containing a training set and a manually verified test set. As shown in Figure~\ref{fig:2}, we design an effective data engine pipeline, which contains video selection and description generation, for automatic data production. We first introduce the video selection part to demonstrate the method of obtaining the target multi-shot videos.

{\noindent\textbf{Video Selection}} Since the multi-shot videos are difficult to distinguish directly and scenes often change with character transitions, we propose using character changes as a proxy. Additionally, We only retain videos up to two minutes in length, focusing on multi-scene content and excluding interference from other issues such as long duration.
 We then employ a coarse-to-fine filtering strategy. In the first stage, a filter based on person instances is employed, using OC-SORT~\cite{cao2023observation}  to perform coarse filtering based on person tracking in order to analyze the trajectories of person IDs. Videos with ID transition exceeding 5 occurrences are retained. In the second stage, we establish two key criteria for the strong close-source video LLM Gemini-1.5-Pro-flash~\cite{team2024gemini} to filter videos: i) Person ID switches within the video, which require transitions in person identities; ii) Clear identification, limiting foreground and background characters to balance scene complexity and computational feasibility.

\noindent\textbf{Video Description Generation} Once the target videos are selected, the next step is to obtain descriptions. However, the widely used dense descriptions cannot meet the requirement of multi-shot videos because they often fail to capture character associations across scenes and tend to overlook short or non-critical segments. To address this, we introduce two new annotations: \textbf{keyframe descriptions} with person IDs and \textbf{character information}, which respectively emphasizing the consistency of characters and their features and actions. 
To obtain additional annotations, we propose an automated framework. As shown in Figure \ref{fig:2}, we first select keyframes using PySceneDetect and then use LLaVA-1.5\cite{liu2024improved} to generate detailed description for the selected keyframes, serving as the initial keyframe description. Then we further leverage Gemini to refine the initial keyframe description with the original video, utilizing its strong ability to detect person ID changes and behavioral shifts (see Supplementary Materials for details). By combining the refined keyframe description and the original video, we ultimately produce the dense caption and character information using Gemini. Thus, we obtain the video description, which encompasses three key elements: keyframe description, character information, and dense captions.


Our video data 
comes from multiple sources, including Kinetics-710 \cite{li2022uniformerv2}, VideoChatGPT \cite{maaz2023video}, VideoChat \cite{li2023videochat}, YouCook2 \cite{li2023videochat}, NExTQA \cite{xiao2021next}, WebVid \cite{yang2021just,bain2021frozen}, and EgoQA \cite{grauman2022ego4d}. After filtering and annotation, each video contains the three elements as shown in Figure \ref{fig:2}, resulting in a descriptive dataset with 23k high-quality video-text pairs (6.7k videos) tailored for multi-shot scences.

\begin{figure*}
\centering
\includegraphics[width=0.9\linewidth]{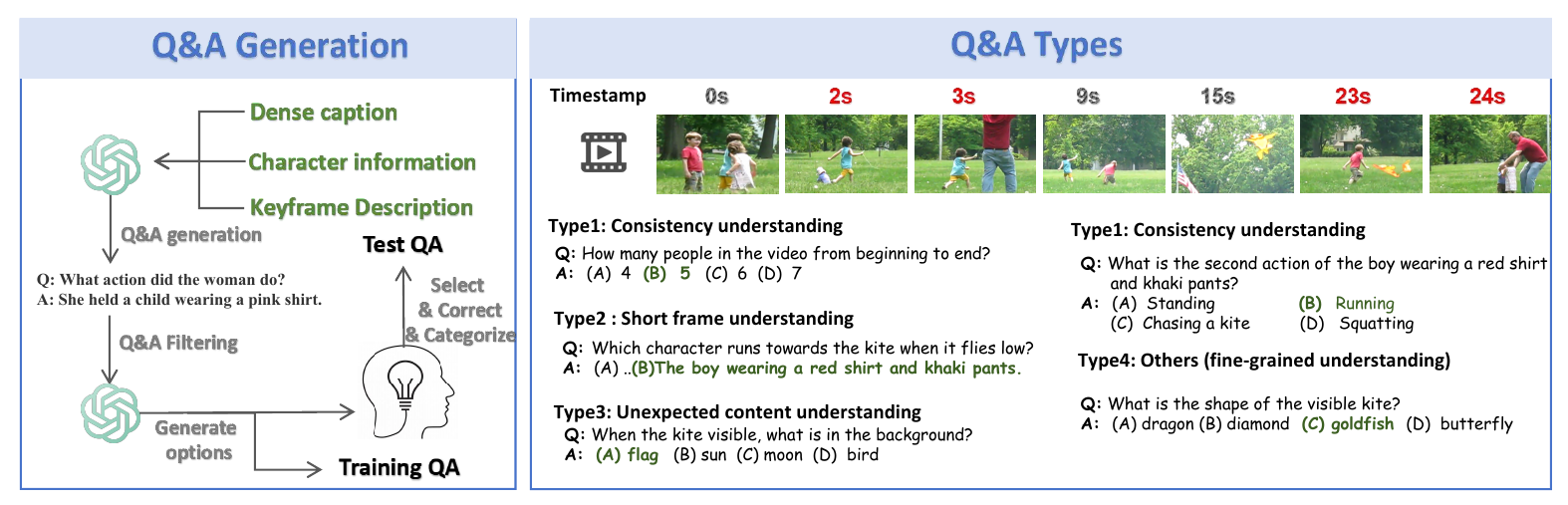}
\caption{\textbf{QA Generation Pipeline and MultiClip-Bench Examples.}  Questions fall into four types: consistency, short-frame, unexpected content, and others. The video shows children flying kites and running. Type 1 involves person re-identification and discontinuous actions. Type 2 focuses on events between seconds 23-24 (short-frame). Type 3 addresses non-critical background details.}
	\label{fig:3}
\end{figure*}

\begin{figure*}[!t]
	\centering
	\includegraphics[width=0.77\linewidth]{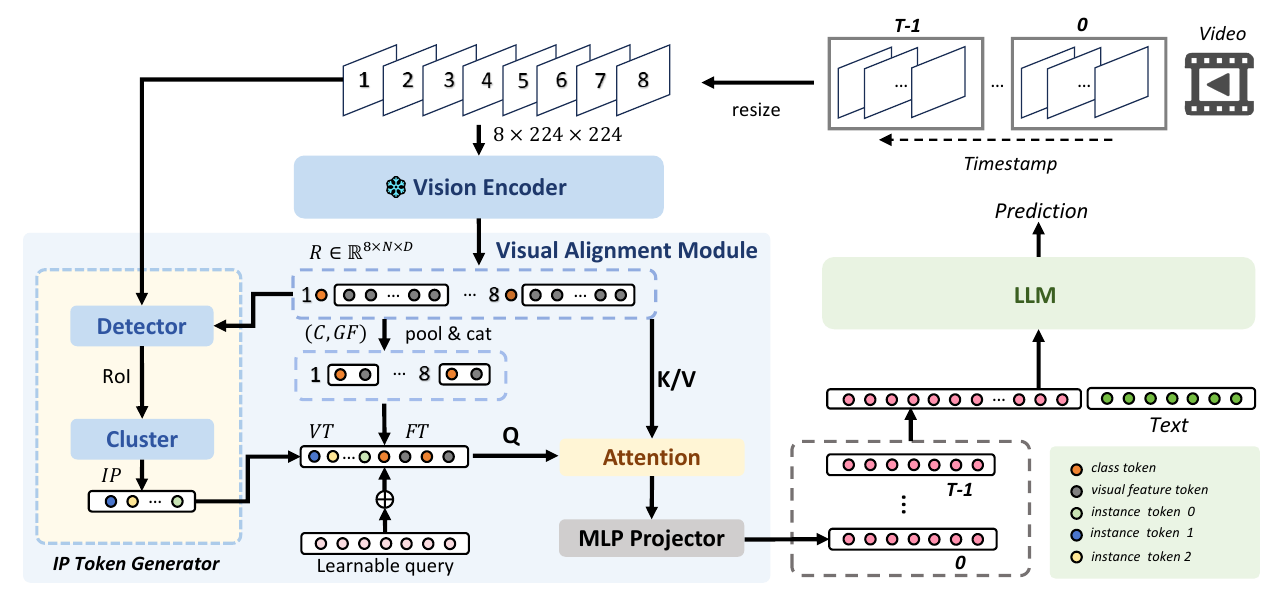}
\caption{\textbf{The pipeline of Our IPFormer-VideoLLM.}
First, video slices are sampled using a sliding window, and a video encoder extracts visual features from each slice. These features are passed to a visual alignment module, where frame-level and instance-level information are extracted separately. Frame-level information includes cls tokens (classification markers for each frame) and avg pooling tokens (via global average pooling), while instance-level information is generated by an IP Token Generator. Next, frame-level and instance-level information are concatenated and injected into a learnable query through addition to optimize it. The optimized query is fed into an attention mechanism for feature fusion, and the fused features are projected through an MLP for text alignment. Finally, the aligned visual features from all video segments are combined with text features and input into a large language model (LLM) for further processing.}

	\label{fig:pipeline}
\end{figure*}
\vspace{-5pt}
\noindent\subsection{QA Pairs of Multi-Shot Scenes}
\label{3.2}

After obtaining the key elements for describing multi-shot videos, the next step is to generate suitable QA pairs for training and testing sets. In addition to enhancing instruction-following capabilities, QA pairs can also improve comprehension abilities. The generation process and examples are shown in the Figure~\ref{fig:3}.

We first explore the challenges of understanding multi-shot videos, which can be generally categorized into three types. QA pairs are generated for each of these challenge types: \textbf{i) Consistency Understanding.} The main focus is on identifying the same characters in non-consecutive keyframes. In particular, this QA type includes but is not limited to, tracking action continuity, such as action counting, temporal action understanding, and character number counting. \textbf{ii) Short Frame Understanding.} This challenge involves addressing modeling deficiencies in capturing the visual cues of keyframes in brief moments. To solve the challenge, questions are posed around these frames, focusing on visual issues such as instantaneous actions, appearance information, and other visually relevant details. \textbf{iii) Unexpected Content Understanding.} The core focus is on handling content that deviates from the main theme, i.e., non-critical unexpected content in multi-shot scenes that does not align with the main video context, which we refer to as ``unexpected content." The questions focus on the individuals and objects appearing in these non-critical, unexpected scenes. These three categories comprehensively address the main challenges of multi-shot transitions.

With the proposed QA types, we employ GPT-4 to create 10 QA pairs per video, focusing on multi-person scenes with shot transitions. Then we pass the generated initial QA pairs to GPT-4 for refinement, where 4 to 6 high-quality QA pairs are retained. Finally, three incorrect options are added to each question, converting them into a multiple-choice format with GPT-4. The videos are sourced from the collection in the previous stage. The training set contains $45.5$k video-text pairs, and the test set contains $2.75$k video-text pairs, with no overlap between the two sets. The manually curated, corrected, and categorized test set includes $800$ consistency understanding questions, $703$ unexpected content understanding questions, $562$ short-frame understanding questions, and $685$ other questions covering fine-grained tasks and general comprehension.

\label{sec:intro}

\section{Visual-Prompt Former for Video LLM}
In addition to the MultiClip-Bench, we propose IPFormer-VideoLLM, which is adapted for multi-shot video understanding. 
This section details the design and structure of IPFormer-VideoLLM. Section \ref{Pipeline of IPFormer-VideoLLM} describes the model architecture. Section \ref{Vision Compressor of IPFormer} discusses the Vision Alignment module. Section \ref{IP Token Generator} introduces the setup of our Visual-prompt tokens.

\subsection{Pipeline of IPFormer-VideoLLM} \label{Pipeline of IPFormer-VideoLLM}
IPFormer-VideoLLM, built on Video-LLaVA, consists of three main modules as illustrated in Figure~\ref{fig:pipeline}: a visual encoder, a visual alignment module (incorporation the IP Token Generator, attention mechanism and MLP), and a large language model (LLM). Due to the fixed number of frame constraint for video encoders, we employ a sliding-window sampling approach. 
Specifically, for an input video, $F$ frames are uniformly extracted. Using a sliding window of $8$ frames, we perform non-overlapping sampling, resulting in $T = \left\lfloor \frac{F}{8} \right\rfloor$ slices. Each slice is processed separately by the visual encoder and the visual alignment module. The processed slices are then concatenated and fed into the LLM along with the text. One significant improvement over previous methods is the visual alignment module, where instance feature tokens generated by the IP Token Generator are used as prompts to guide the attention mechanism in aggregating instance features.

\subsection{Vision Alignment Module of IPFormer} \label{Vision Compressor of IPFormer}


\textbf{Motivation: }Based on our research on data, we assume that visual alignment modules for multi-shot video understanding should meet two requirements: i) Effective compression capability for vision token, allowing them to handle many frames. ii) Instance awareness, enabling the capture of important instance information across scenes. However, current visual alignment modules, whether full projection or compression before projection, encode instance features in a discrete or lossy manner, inevitably losing instance-level information. Inspired by these requirements, we propose an instance-based compressor as a visual alignment module, which reduces key feature loss while compressing visual tokens.


\noindent\textbf{Method: }To enable the attention mechanism to capture more critical information, 
we introduce ``anchors"~\cite{ren2015faster,meng2021conditional} during the initialization of learnable query tokens inspired by Conditional-DETR~\cite{meng2021conditional}.
These anchors help the learnable tokens extract relevant information from frames, 
thereby better preserving key information. Thus, we input both instance-level and frame-level information as ``anchors" into the learnable query tokens. 
Specifically, 
prior to encoding, the video is pre-processed into images of size $224 \times 224$ for 8 frames. The video encoder then processes these 8 frames into $\mathbf{R} \in \mathbb{R}^{8 \times N \times D}$, where $N$ is the number of tokens per frame, and $D$ is the vector dimension. Here, $N = 256 + 1$, where $256$ denotes the number of image tokens per frame, called $\mathbf{N}_I \in \mathbb{R}^{256 \times D}$, and $1$ is the class token per frame, referred to as $\mathbf{C} \in \mathbb{R}^{1 \times D}$. Next, global average pooling in performed on $\mathbf{N}_I$ to obtain the global feature of each frame, $\mathbf{GF} \in \mathbb{R}^{1 \times D}$.  We use $\mathbf{GF}$ and $\mathbf{C}$ as global information for each frame (frame-level guidance). To enhance frame information learning,  $\mathbf{GF}$ and $\mathbf{C}$ are repeated each $X$ times ($X$=$5$), then concatenated to obtain the frame token for each frame: $\mathbf{FT} = (\mathbf{C} \times X, \mathbf{GF} \times X) \in \mathbb{R}^{(X \times 2) \times D}$. 
In addition, we also obtain instance features through the IP Token Generator. The instance prompt tokens for this segment are represented as $\mathbf{IP} \in \mathbb{R}^{V \times D}$, where $V$ is the number of instance tokens. We concatenate $\mathbf{FT}$ of $8$ frames and $\mathbf{IP}$, resulting in the visual tokens for 8 frames as $\mathbf{VT} \in \mathbb{R}^{8 \times (X \times 2 ) \times D+ V\times D}$. Finally, $\mathbf{VT}$ is added to the learnable queries to optimize them, guiding the learnable queries to aggregate the video features in the cross-attention mechanism, thereby better capturing relevant instance or other useful information. The output of the attention is then projected using an MLP to align with the textual space.


\subsection{IP Token Generator} \label{IP Token Generator}
The IP Token Generator generates instance tokens as prompts. Specifically, we employ a category-agnostic detector \cite{zhu2020deformable}, with initialization parameters derived from~\cite{ma2024groma}. The detector's original classification head is replaced by a binary classifier to score region proposals based on their localization quality. This process yields a series of bounding boxes, which are subsequently processed through Non-Maximum Suppression~\cite{neubeck2006efficient}. 
For each frame, we retain only $M$ (fewer than $10$) candidate boxes, truncating any excess and padding with zeros when necessary. 
These candidate boxes are processed using global RoI pooling on the visual feature maps of each frame to obtain instance features. Subsequently, we compute the cosine similarity between the instance features across all frames within a video slice. We set a threshold ($0.9$) 
 and adopt an iterative grouping strategy. Starting from the first ungrouped instance, we traverse all remaining ungrouped instances,  gathering those with similarities exceeding the threshold to form a group. Subsequent rounds of traversal are performed until no further groupings can be formed. Any residual instances failing to meet the similarity criteria ultimately constitute independent groups. For each group, the channel-wise average is computed to obtain aggregated instance features, referred to as the Instance Prompt. The number of Instance tokens per slice is limited to a maximum of $V$ ($80$), with zeros padded when fewer than $80$ tokens are present. The purpose of clustering is to balance the number of instances, preventing instances that appear frequently from overwhelming sparse ones. This approach is particularly beneficial for short frame understanding.

\section{Experiments}

\begin{table}[htbp]
\centering
\scriptsize
\setlength{\tabcolsep}{2pt} 
\renewcommand{\arraystretch}{1.0}
\resizebox{\linewidth}{!}{ 
\begin{tabular}{@{}lccccc@{}}
\hline
Method      & Frame & Token & Train Time & FPS & Act-Net \\
\hline
\rowcolor{blue!5}Video-LLaVA & 8 & 2056 & 41h & 1.1 & 45.3 \\
\hline
Ours (8f) & 8 & 160 & 10h & 3.3 & 45.5 \\
Ours (16f) & 16 & 320 & - & 2.0 & 46.1 \\
\hline
\end{tabular}
}
\caption{Efficiency comparison between Video-LLaVA and ours. The highlighted row in \textcolor{blue!30}{blue} is our baseline model.}\label{tab:1}
\end{table}

 \begin{table*}\small
 \centering
 \renewcommand{\arraystretch}{1}
\addtolength{\tabcolsep}{1pt}
 {
\begin{tabular}{@{}l|c|c|c|cccc|c@{}}
\hline
Method & LLM size &Frames& Tokens &Consistency uds &Short uds&Unexpected content&Others  &Mean \\ \hline
\rowcolor{blue!5} Video-LLaVA~\cite{lin2023video} &7B & 8 &2056 &32.9&40.0 & 46.2&60.7& 44.5 \\
VideoChat2~\cite{li2024mvbench}  &7B & 16&96& 33.4& 32.3 &47.6  & 58.1& 42.9\\
ST-LLaVA~\cite{qu2024ts}  &7B  & 100 &2304&31.1&38.8&48.3  &56.8&43.5 \\

PLLaVA~\cite{xu2024pllava} &7B  & 16 &2304&36.5&43.0 &55.9 &63.2&49.4\\ 
LLaVA-MINI~\cite{zhang2025llava}&8B   & 64 &64&26.8& 40.9& 53.2&58.6&44.9\\

 \hline
 \textbf{Ours} (w/o MultiClip) & 7B &  16&320 &37.7&43.1& 48.4&61.9&48.1\\ 
 \textbf{Ours} & 7B&  16&320&46.1 &53.6&60.4&66.3&57.0\\ 
  \textcolor{DarkGreen}{\textbf{$\Delta Acc$}.}
 & -&  -&- &\textcolor{DarkGreen}{\textbf{+8.4}}&\textcolor{DarkGreen}{\textbf{+10.5}}& \textcolor{DarkGreen}{\textbf{+12.0}}&  \textcolor{DarkGreen}{\textbf{+4.4}}& \textcolor{DarkGreen}{\textbf{+8.9}}  \\ 
 \textbf{Ours} &7B &  48&960&\textbf{48.6}& \textbf{56.7}& \textbf{62.5}&68.5&\textbf{58.8}\\ 
  \hline
  GPT-4o-mini~\cite{abacha2024medec}   &8B &16&Unk & 44.7&44.6 &55.3 &59.1 & 50.9\\
GPT-4V\cite{yang2023dawn}   &Unk &8&Unk &34.6&31.5 &45.3  &52.4 & 41.8 \\
  Qwen2.5-vl~\cite{bai2025qwen2}    &72B&16&Unk &43.5&\textbf{58.3} & \textbf{62.4} &72.8 &58.5\\ 
Gemini-1.5-pro~\cite{team2024gemini}  &Unk & 16&Unk&\textbf{53.4} & 52.5& 62.0 &\textbf{74.4} & \textbf{60.5}\\

\hline
\end{tabular}
}
\caption{Comparison with other models on MultiClip-Bench. Best performance is \textbf{boldface}. The highlight in \textcolor{blue!30}{blue} is our baseline model.}
\label{tab:2}%
\end{table*}

\begin{table}[t]
    \centering
    \footnotesize
    \resizebox{\linewidth}{!}{
    \begin{tabular}{@{}lc|ccc@{}}
        \hline
        \multirow{2}{*}{Method}  & \multirow{2}{*}{LLM Size} 
        & \multicolumn{3}{c}{\textbf{QA Metrics}} \\
        \cline{3-5}
        & & MSVD & MSRVTT & ActivityNet \\ 
        \hline
        FrozenBiLM~\cite{yang2022zero}  & 1.3B & 33.8 & 16.7 & 25.9 \\
        Video-LLaMA~\cite{zhang2023video} & 7B & 51.6 & 29.6 & 12.4 \\
        LLama-Adapter~\cite{zhang2023llama}  & 7B & 54.9 & 43.8 & 34.2 \\
        Video-ChatGPT~\cite{maaz2023video}& 7B & 64.9 & 49.3 & 35.3 \\
        VideoChat~\cite{li2023videochat} & 7B & 56.3 & 45.0 & 26.5 \\
        \rowcolor{blue!5}  Video-LLaVA~\cite{lin2023video}&7B & 70.7 & 59.2 & 45.3 \\
        Chat-UniVL~\cite{jin2024chat}  &7B & 65.5 & 54.6 & 45.8 \\
        MovieChat~\cite{song2024moviechat} & 7B & 75.2 & 52.7 & 45.7 \\
        VideoChat2~\cite{liu2024mmbench} &7B & 70.0 & 54.1 & 49.1 \\
        Vista-LLama~\cite{ma2023vista} & 7B & 65.3 & 60.5 & 48.3 \\
        LLAMA-VID\cite{li2024llama} & 13B & 70.0 & 58.9 & 47.5 \\
        ST-LLM~\cite{liu2024st}& 7B & \textbf{74.6} & \textbf{63.2} & \textbf{50.9} \\
        \hline
        Ours & 7B & 73.8 & \textbf{63.2} & 50.1 \\
        \hline
    \end{tabular}
    }
    \caption{Quantitative comparison on video question-answering.}
    \label{tab:results_video_qa}
\end{table}

\begin{table}[h]
\footnotesize
\addtolength{\tabcolsep}{4pt}
\resizebox{\linewidth}{!}
{
\begin{tabular}{@{}l|c|c|c@{}}

\hline
\textbf{Method} & \textbf{Vision Encoder} & \textbf{LLM Size} & \textbf{Avg.} \\ \hline
Video-LLaMA \cite{zhang2023video} & CLIP-G & 7B & 34.7 \\
LLaMA-Adapter \cite{zhang2023llama} & ViT-B & 7B & 31.7 \\
Video-ChatGPT \cite{maaz2023video} & ViT-L & 7B &  32.7 \\
VideoChat\cite{li2023videochat}  & CLIP-G & 7B &35.5 \\
\rowcolor{blue!5}  Video-LLaVA\cite{lin2023video}  &ViT-L&7B& 42.2 \\
VideoChat2 \cite{li2024mvbench} & UMT-L & 7B & \textbf{51.1} \\
PLLaVA  \cite{xu2024pllava} & ViT-L & 7B & 46.5 \\
LLaVA-MINI \cite{zhang2025llava}  & ViT-L&8B & 44.5 \\

\hline
Ours     &ViT-L&7B & 48.3 \\ 
\hline
\end{tabular}
}
\caption{Quantitative comparison on MVBench.}
\label{tab:mvbech}
\end{table}



\begin{table}\footnotesize
    \centering
    \addtolength{\tabcolsep}{8pt}
    
    \begin{tabular}{@{}l|c|c|c@{}}
        \hline
        Method  & NExT-QA & Egoschema &IntentQA \\
        \hline
        \rowcolor{blue!5} Video-LLaVA \cite{lin2023video}   & 60.5 & 37.0& -\\
        Video-LLaMA2 \cite{cheng2024videollama}   & - & 51.7 &- \\
        MovieChat+ \cite{song2024moviechat}   & 54.8 & 56.4 &- \\
        Vista-LLaMA \cite{ma2023vista}   & 60.7 & - & - \\
        DeepStack-L \cite{meng2025deepstack}   & 61.0 & 38.4  &-\\
        M$^3$ \cite{cai2024matryoshka}  & 63.1 & 36.8  &58.8\\
        IG-VLM \cite{kim2024image}   & 63.1 & 35.5  &60.3\\
        SF-LLaVA \cite{xu2024slowfast}  & 62.4 & 47.6 & 60.1\\
        TS-LLaVA \cite{qu2024ts}  & 66.5 & \textbf{50.2}& 61.7\\
        \hline
        ours   &  \textbf{70.6}& 50.0 & \textbf{75.3}\\
        \hline
    \end{tabular}
    \caption{Quantitative comparison on Multiple-choice question-answering in specific fields.}
    \label{tab:multiple_choice}
    \label{tab:results}
\end{table}

\subsection{Implementation Details}

\noindent \textbf{Model Details: }To validate the effectiveness of the proposed modules, we use Video-LLaVA~\cite{lin2023video} as the baseline, maintaining the same training settings to ensure the only difference are in the model modifications. Specifically, visual encoder from LanguageBind~\cite{zhu2023languagebind} is used, initialized from OpenCLIP-L/14~\cite{radford2021learning}, with a resolution of 224×224. For large language model, we use the LLM in LanguageBind initialized from Vicuna-7B v1.5~\cite{zheng2023judging}. The text tokenizer is from LLaMA~\cite{touvron2023llama} with about 32,000 classes. 

Regarding our design, in addition to the two-layer MLP full projection with GeLU~\cite{hendrycks2016gaussian} for the visual compressor, we adopt the approach of BLIP2~\cite{li2023blip} and VideoChat2~\cite{li2024mvbench}, implementing an attention parameter compression module using pre-trained BERTbase~\cite{devlin2019bert}. For the external detector, we adopt the Region Proposer parameters from Groma~\cite{ma2024groma}, implementing a modified binary classifier based on Deformable DETR~\cite{zhu2020deformable}.

\noindent\textbf{Training Details: }Our training process follows Video-LLaVA~\cite{lin2023video} and is divided into two stages: modality alignment pretraining and instruction fine-tuning. 

During \textit{modality alignment pretraining}, we follow Video-LLaVA using $558$K image-text pairs from LAION-CC-SBU~\cite{liu2023visual,sharma2018conceptual} and $702$K video-text pairs from Valley~\cite{luo2023valley}. We train the visual alignment module for one epoch (batch size 256) using AdamW optimizer (lr=$1e^{-3}$, warmup=$0.03$), while freezing other parameters. Each video contains $8$ frames at a $224 \times 224$ resolution.

For the \textit{instruction tuning} stage, we use the same base data as Video-LLaVA ($665$K image-text pairs from LLaVA-v1.5 and $100$K video-text pairs from Video-ChatGPT). To further enhance performance, we additionally selected $326$K instruction pairs from VideoChat2~\cite{li2024mvbench}, $27$K data from Perception~\cite{patraucean2023perception} and STAR~\cite{wu2024star} datasets, and $68$K video-text pairs from our MultiClip dataset. We also increase the number of sampled frames from 8 to 16 using a sliding window approach to benefit multi-shot scenes. The training settings remain consistent with Video-LLaVA: keeping video encoders frozen while fine-tuning all LLM parameters with a batch size of 128.



\subsection{Efficiency of IPFormer-VideoLLM}


To address the computational bottleneck in Video-LLaVA's visual processing, we propose replacing its fully connected layers with an attention-based visual compressor. We validate its efficiency through comprehensive timing experiments, as shown in Table~\ref{tab:1}. The results demonstrate that this modification significantly reduces visual tokens while maintaining model performance. Specifically, under the same hardware configuration (8×A100-80G GPUs) and experimental settings, our approach reduces the first-stage training time to 1/4 of the original duration, shortening the entire training process from 3–4 days to 1–2 days. For inference evaluation (conducted on a single 3090Ti), while the timing results in Table~\ref{tab:1} exclude box detection time (as we use pre-computed box labels), our model still achieves 1.4 FPS with 16-frame processing even when including real-time box detection, demonstrating strong computational efficiency. In subsequent experiments, unless otherwise specified, inference is performed with \textbf{$16$} frames by default.

\subsection{Result on MultiClip-Bench}

We compare our model with existing open-source video multimodal models on our self-constructed MultiClip-Bench dataset, with results shown in Table~\ref{tab:2}. By comparing Video-LLaVA and Ours (w/o MultiClip), it is obvious that our model design leads to a significant improvement (\textcolor{DarkGreen}{\textbf{+3.6\%}}) 
When incorporating our proposed training dataset, performance improves further by \textcolor{DarkGreen}{\textbf{+8.9\%}}. Additionally, increasing the frame number can help understanding.

We also test several state-of-the-art large multimodal models (most of them are closed-source) using a multi-image evaluation approach. Among these, Qwen2.5-VL~\cite{bai2025qwen2} and Gemini-1.5-pro~\cite{team2024gemini} each demonstrate their strengths in different categories.



\subsection{Results on other video benchmarks}

Besides the multi-shot scenes, we also evaluate our model on various video benchmarks to demonstrate its versatility. For open-ended Video Question Answering tasks, including MSVD-QA~\cite{chen2011collecting}, MSRVTT-QA~\cite{xu2016msr}, and ActivityNet-QA~\cite{yu2019activitynet}, where answers are typically short phrases, we follow Video-LLaVA to use GPT-3.5-turbo for accuracy evaluation, with results shown in Table~\ref{tab:results_video_qa}. We also assess performance on MVBench, a comprehensive multi-choice benchmark comprising 20 tasks that require detailed temporal understanding of videos, as presented in Table~\ref{tab:mvbech} (see supplementary materials for details). Furthermore, we validate on more challenging multi-choice benchmarks, including NExT-QA~\cite{xiao2021next}, EgoSchema~\cite{grauman2022ego4d}, and IntentQA~\cite{li2023intentqa}, with results shown in Table~\ref{tab:multiple_choice}. Through combined improvements in data and model architecture, our approach significantly outperforms other methods, particularly in NExT-QA and IntentQA. This superior performance can be attributed to the effective coverage of multi-shot scenes and person-centric questions within our dataset.

\subsection{Ablation study}
This section, we analyze the effect of specific model design and dataset, including the comparison of our visual compression module with other visual alignment methods, the specific design of visual compression module, and the impact of the proposed dataset and frame-increasing strategy.

\begin{table}
\footnotesize
\addtolength{\tabcolsep}{0pt}
\centering
\begin{tabular}{@{}l|c|cc|cc@{}}
\toprule
  & \multirow{2}{*}{Frame Query} &  \multicolumn{2}{c|}{Instance Query}&\multirow{2}{*}{ActivityNet}&\multirow{2}{*}{MultiClip}\\
& & no-cluster& cluster &   & 
\\
\midrule
S$1$ & & & &40.9 &42.3\\
S$2$ &\ding{52} & & &42.6  &43.1\\
\midrule
S$3$& \ding{52} & \ding{52}&&43.6 &44.6\\
S$4$& \ding{52} &&\ding{52} &45.5&46.2 \\
\bottomrule
\end{tabular}
\caption{Design of visual compression module.}
\label{table:8}
\end{table}

\begin{table}\footnotesize
\centering
\renewcommand\arraystretch{1.1}
\addtolength{\tabcolsep}{-3pt}
\resizebox{\linewidth}{!}
{
\begin{tabular}{cc|ccccc}

\hline
  \multicolumn{2}{c|}{Visual Alignment Module} & \multicolumn{4}{c}{Metric} \\ \cline{1-2} 
 structure & Token &  Activity   &MSRVTT&MVBench&MultiClip   \\ \hline
Full-projection  & 8*256 &45.3 &60.2 &42.2&44.5\\
\hline
Avg-pooling (2*2)  & 8*64 &45.0 &- &41.7&-\\
QFormer (S1)  & 160& 40.9 &- &40.1&42.3\\
 \hline
IPFormer (S4) & 8*10+80 &45.5 &61.8 &42.3&46.2\\
 \hline
\end{tabular}
}
\caption{Comparison of different visual alignment methods. Avg-pooling and QFormer are common visual compression techniques.}\label{tab:5}
\end{table}

\begin{table}\footnotesize
\centering
\renewcommand\arraystretch{1}
\addtolength{\tabcolsep}{2pt}

\resizebox{\linewidth}{!}
{
\begin{tabular}{@{}ccccc@{}}
\hline
 &Data source  &ActivityNet& MVBench  &MuitiClip
\\ \hline
D$1$&  Video-LLaVA&45.8&43.7&47.1\\
D$2$& +VideoChat2(353k)& 48.6&47.5& 48.1\\
D$3$ &+ MultiClip(68k)&50.1  &48.3& 57.0\\
\hline
\end{tabular}
}
\caption{The effect of training dataset.}\label{tab:4}
\end{table}

\noindent\textbf{Design of the Visual Compression Module: }Here we verify the effect of each design aspect within our visual compression module, with results shown in Table~\ref{table:8}. Comparison of S1 and S2 shows that incorporation of frame queries leads to $1.7\%$ performance gain compared to the original learnable queries in Q-Former. Additionally, injecting instance to queries can further improve the performance, but without clustering, the gain is limited. With clustering of instance tokens, performance increases to $46.2$ on MultiClip.


\noindent\textbf{Comparison of Different Visual Alignment Methods:} 
We compare our visual compression module with other common methods for visual alignment, with results shown in Table~\ref{tab:5}. Full projection, commonly used in Video-LLaVA, does not compress the tokens and thus minimizes information loss. Average pooling, adopted by PLLaVA, LLaVA-Next~\cite{li2024llava}), has a low compression rate, resulting in only marginal performance drops. In contrast, the original Q-Former (S1 in Table~\ref{table:8}) results in significant information loss, which largely impacts the performance. However, with our query design (S4 in Table~\ref{table:8}), the Q-Former architecture can surpass the full-projection while compressing tokens.

\noindent \textbf{Effect of Dataset:}
To validate the effectiveness of our proposed dataset, we conduct experiments to evaluate its impact on model performance. As shown in Table~\ref{tab:4}, the limited instruction data in 
VideoChatGPT~\cite{maaz2023video} used by Video-LLaVA are insufficient to meet the needs for fine-grained and consistent understanding in real scenarios. By incorporating VideoChat2 instruction data and our proposed dataset, we observe substantial performance improvements, which demonstrate that our keyframe annotation strategy significantly enhances model capabilities. Notably, with only 68k training samples, we achieve significant improvements on the MultiClip benchmark.  We also test the performance changes when using different numbers of frames for training and inference, demonstrating the scalability of the model. 
 (Per supplementary materials)




\section{Conclusion}


Recognizing a gap not previously addressed in the handling of multi-shot videos with VideoLLMs, we identify the deficiencies and their underlying causes. This lead us to propose proprietary datasets and a novel design. By developing an automated video annotation pipeline, we tackle data scarcity in multi-shot scenarios. Furthermore, we introduce an instance-based visual compression module for multi-scene challenges. Our dataset and model significantly enhance multi-scene video understanding and demonstrate unique advantages across various video benchmarks.

{
    \small
    \bibliographystyle{ieeenat_fullname}
    \bibliography{main}
}

\end{document}